\pdfoutput=1

\documentclass[11pt]{article}
\usepackage{authblk}
\usepackage{graphicx}
\usepackage{amsmath}
\DeclareMathOperator*{\argmax}{arg\,max}
\usepackage{url}

\usepackage{emnlp2021}

\usepackage{times}
\usepackage{latexsym}

\usepackage[T1]{fontenc}

\usepackage[utf8]{inputenc}

\usepackage{microtype}

\title{Medical Question Understanding and Answering \\ with Knowledge Grounding and Semantic Self-Supervision}

\author[1]{\bf Khalil Mrini}
\author[1]{\bf Harpreet Singh}
\author[2]{\bf Franck Dernoncourt}
\author[2]{\bf Seunghyun Yoon}
\author[2]{\\ \bf Trung Bui}
\author[2]{\bf Walter Chang}
\author[1]{\bf Emilia Farcas}
\author[1]{\bf Ndapa Nakashole}
\affil[1]{University of California, San Diego, La Jolla, CA 92093 \protect\\ \small{\texttt{\{khalil, h1singh, efarcas, nnakashole\}@ucsd.edu}}}
\affil[2]{Adobe Research, San Jose, CA 95110 \protect\\ \small{\texttt{\{franck.dernoncourt, syoon, bui, wachang\}@adobe.com}}}

\begin{document}
\maketitle

\begin{abstract}
Current medical question answering systems have difficulty processing long, detailed and informally worded questions submitted by patients, called Consumer Health Questions (CHQs). To address this issue, we introduce a medical question understanding and answering system with knowledge grounding and semantic self-supervision. Our system is a pipeline that first summarizes a long, medical, user-written question, using a supervised summarization loss. Then, our system performs a two-step retrieval to return answers. The system first matches the summarized user question with an FAQ from a trusted medical knowledge base, and then retrieves a fixed number of relevant sentences from the corresponding answer document. In the absence of labels for question matching or answer relevance, we design 3 novel, self-supervised and semantically-guided losses. We evaluate our model against two strong retrieval-based question answering baselines. Evaluators ask their own questions and rate the answers retrieved by our baselines and own system according to their relevance. They find that our system retrieves more relevant answers, while achieving speeds 20 times faster. Our self-supervised losses also help the summarizer achieve higher scores in ROUGE, as well as in human evaluation metrics. We release our code to encourage further research.\footnote{Link: \url{https://github.com/KhalilMrini/Medical-Question-Answering}}
\end{abstract}

\section{Introduction}

\begin{figure}
    \centering
    \includegraphics[width=\columnwidth]{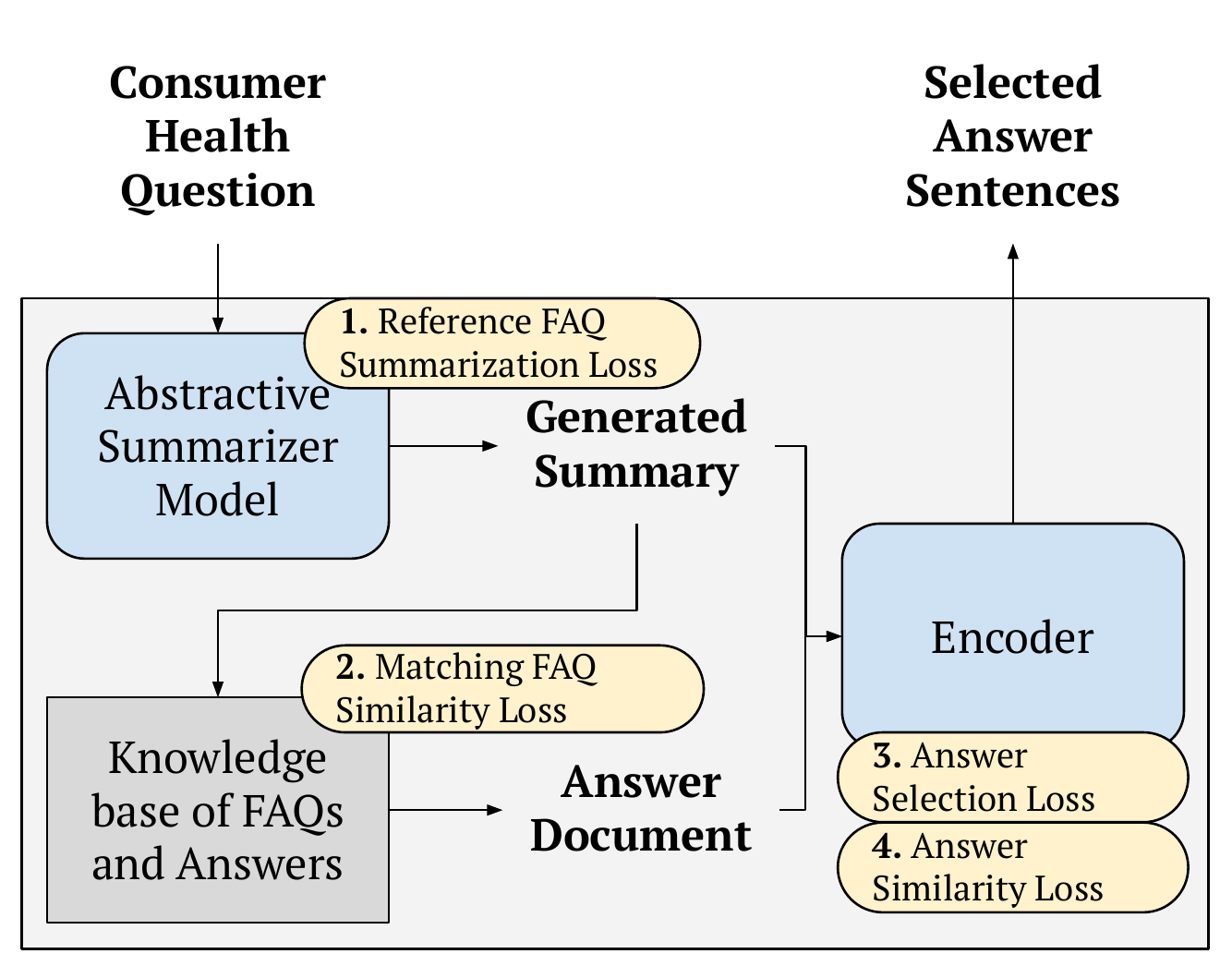}
    \caption{Overview of our proposed Consumer Health Question Understanding and Answering model. The input is a user question, called \textit{Consumer Heath Question} (CHQ). The goal is to match the CHQ to relevant answer sentences associated with a \textit{Frequently Asked Question} (FAQ) from a medical knowledge base.}
    \label{summary}
    
\end{figure}

\textbf{Motivation.} Users of medical question answering systems often write long questions, called Consumer Health Questions (CHQs). Several aspects of CHQs hinder the capacity of current question answering (QA) systems to process them: long medical questions may contain peripheral information like patient history \cite{roberts2016interactive} that are not necessary to retrieve relevant answers. Consumer health questions may also use a distinct vocabulary from the one used by medical providers to describe the same health concepts \cite{abacha2019summarization}.

A growing number of approaches attempt to enhance the processing of consumer health questions -- or medical question understanding. These approaches include query relaxation \cite{abacha2015means, lei2020expanding}, question entailment \cite{abacha2016recognizing, abacha2019question, agrawal2019ars_nitk}, question summarization \cite{abacha2019summarization}, and question similarity \cite{abacha2017nlm_nih, yan2018medical}. 

However, the above medical question understanding approaches stop short of retrieving answers after processing consumer health questions. The Medical Question Answering Task
at TREC 2017 LiveQA \cite{abacha2017overview} attempts to fill the gap by proposing the task of Consumer Health Question Answering. The goal is to retrieve relevant answers obtained using online search for the corresponding CHQ. As part of their participation in this task, \citet{yang2017cmu} find that online search engine queries introduce noise in performance, and that even collected and curated medical knowledge available offline can fare better.

\noindent \textbf{Contributions.} To enable the use of a curated medical knowledge base for answering long user questions, we introduce a novel, knowledge-grounded and semantically self-supervised system for Consumer Health Question Understanding and Answering (CHQUA). We tackle a challenging aspect of CHQUA: providing answers when no relevance labels are available. Our contributions are as follows:

\textbf{(1)} We propose an end-to-end pipeline, as shown in Figure~\ref{summary}, that takes as input a consumer health question, and trains a summarizer model to generate a short, formally worded question. We optimize a summarization training objective using the medical question summarization datasets.

\textbf{(2)} The medical knowledge base we use is separate from the question summarization datasets, and therefore we have no labels to indicate which knowledge base question matches a given consumer health question. We design a novel, semantically-guided self-supervised loss function to ground the generated summary with knowledge base FAQs, using semantic similarity as proxy to question matching. The Matching FAQ similarity loss helps the encoder pick the most semantically similar knowledge base question.

\textbf{(3)} The large medical knowledge base we use has no answer sentence relevance labels. We adapt to this scenario by designing two complementary self-supervised losses on the same encoder, and by considering semantic similarity as a proxy to relevance. The Answer Similarity loss pushes the model to distinguish between relevant and irrelevant answer sentences, whereas the Answer Selection loss works in a complementary way to push the model to select a given number of sentences.

Finally, we conduct an evaluation to compare the relevance of our system with two strong baselines of retrieval-based question answering. We ask evaluators to ask their own questions, and then perform a blind evaluation of the retrieved answers by each system. Seven evaluators find that our system retrieves more relevant answers compared to the two baselines, while achieving significantly faster processing speeds. We also find that the self-supervised losses help achieve better scores in ROUGE and human evaluation metrics. However, we find that the task remains challenging, with room for improvement. We release our code, model, and matched datasets to encourage further research in consumer health question understanding and answering.

\section{Related Work}

\noindent \textbf{Consumer Health Question Answering.} \citet{abacha2017overview} introduce the Medical QA shared task at TREC 2017 LiveQA, where the goal is to develop a consumer health question answering system. The training data is comprised of question-answer pairs. The questions are informally worded CHQs received by the U.S. National Library of Medicine (NLM). The answers are formally worded and come from websites of the U.S. National Institutes of Health or manually collected by librarians. The evaluation scores are given by humans, using a test set of CHQs and reference answers.



Many participating teams adopt a question matching approach, and train their models on question similarity datasets like the Quora question pair dataset \cite{iyer2017first}, or other datasets collected from community question answering websites. TODO \cite{mrini-etal-2021-joint}

In the MEDIQA 2019 Shared task, \citet{abacha2019summarization} introduce a differently defined consumer health question answering task. Here, the goal is to rank a given list of answers according to their relevance with regard to a CHQ. \citet{he2020infusing} introduce a new disease knowledge infusion training procedure for BERT \cite{devlin2019bert} that scores well in this task.

\noindent \textbf{Medical Question Answering.} Medical QA approaches include translating questions to SPARQL queries \cite{10.1145/2110363.2110372}, semantic similarity between questions and candidate answers \cite{hao2019exploiting}, knowledge representations \cite{terol2007knowledge, 10.1145/3106745}, ranking candidate answers \cite{abacha2017overview, abacha2019overview}, summarization of questions and/or answers \cite{MEDIQA2021, mrini-etal-2021-ucsd, mrini-etal-2021-joint,  mrini-etal-2021-gradually}, and medical entity linking \cite{basaldella2020cometa, mrini-etal-2022-detection}.

There is a variety of definitions for the task of medical QA and related sub-tasks in the literature. \citet{hao2019exploiting} define medical QA as the task of finding the correct answer from a set of candidates and a body of evidence documents. They propose to work on two datasets: the National Medical Licensing Examination of China (NMLEC) \cite{shen2020generation}, and Clinical Diagnosis based on Electronic Medical Records (CD-EMR), where the goal is to predict the correct diagnosis based on patient history.

\citet{sharma2018bioama} propose to tackle three kinds of medical questions found in the BioASQ challenge \cite{bioasq}: factoid questions where answers are single entities, list-type questions where answers are a set of entities, and yes/no questions.

\noindent \textbf{Retrieval-based Question Answering.} Recent methods for retrieval-based QA systems use contextual text embeddings to evaluate a candidate answer's relevance to a given question.

\citet{tay2018multi} propose to use Multi-Cast Attention Networks (MCAN), a new attention mechanism, to model question-answer pairs. 

\citet{mrini-etal-2021-recursive} introduce a recursive, tree-structured model that models sentences according to their syntactic tree. Their results show that tree structure sets a new state of the art in conventional, formally worded QA benchmarks like TrecQA and WikiQA \cite{yang2015wikiqa}, but does not fare well in informally worded, user-written datasets.

\citet{karpukhin-etal-2020-dense} introduce Dense Passage Retrieval (DPR): a dual-encoder based on BERT \cite{devlin2019bert}, that predicts relevance scores of passages with regard to a question. DPR encoders are trained on the relevance of passages from datasets containing such labels, using a supervised negative log-likelihood loss based on the semantic similarity of questions and relevant passages.

\citet{mao2021generation} modify the \textit{query} part of retrieval-based QA: they propose to use language models to generate context for queries. They then feed the extended queries to retrieval systems, such as DPR or BM-25.

\section{Problem Definition}

\begin{figure*}
    \centering
    \includegraphics[width=410pt]{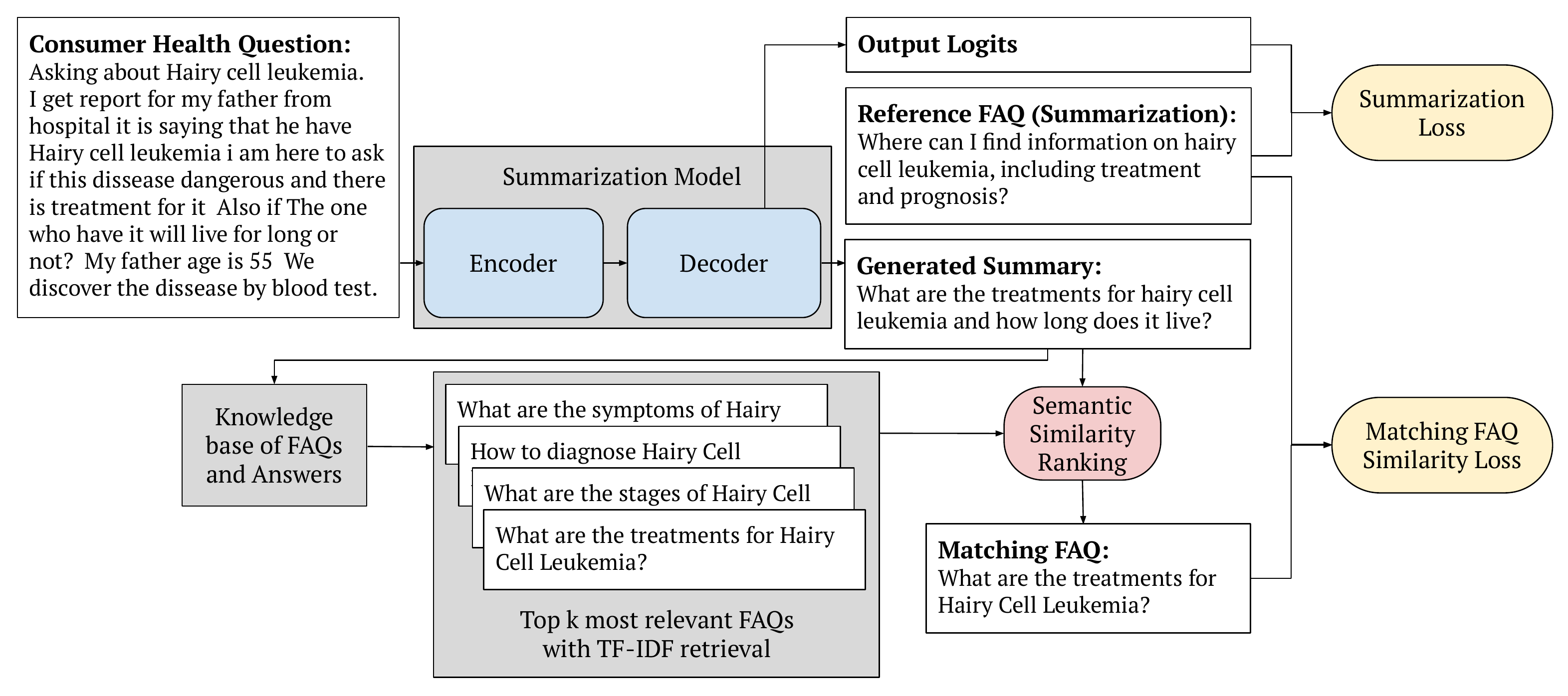}
    \caption{The Consumer Health Question (user question) is first summarized, and we then retrieve a relevant question from the knowledge base using the generated summary. The \textbf{top} half of the figure illustrates the first step: question understanding through summarization (\S \ref{question_und}). The \textbf{bottom} half of the figure illustrates the second step: question matching through self-supervised knowledge grounding (\S \ref{question_ret}).}
    \label{knowledge}
    
\end{figure*}

We define knowledge-grounded Consumer Health Question Understanding and Answering (CHQUA) as the problem of retrieving a fixed number of answer sentences from a medical knowledge base that are the most relevant given a long and informal user question -- called a Consumer Health Question (CHQ). There are three steps in CHQUA: question summarization, matching the summarized user question with a relevant FAQ from the knowledge base, and retrieval of the relevant answer sentences from the corresponding answer document.

Knowledge-grounded CHQUA is comprised of three elements used for training. First, the CHQ is the input of the task. Second, the Reference FAQ (Frequently Asked Question) is the golden or expert-written summary corresponding to the CHQ. Whereas the CHQ is a long and informally worded question, the reference FAQ is the corresponding short, one-sentence, formally worded question. At inference time, the reference FAQ is not available, and we will therefore use a summary generated by the model. Third, the medical knowledge base is comprised of FAQs, where each FAQ has a corresponding answer document with at least one sentence. FAQs in the knowledge base are also short, one-sentence, formally worded questions.

The goal of knowledge-grounded CHQUA is to find a set $\mathcal{R}$ of $n$ relevant answer sentences, from a document comprised of answer sentences $\mathcal{A}_i$, such that $\mathcal{A}_i$ corresponds to question $\mathrm{q}_i$ from the knowledge base. We call $\mathrm{q}_i$ the retrieved or matching FAQ, such that $\mathrm{q}_i$ is the most similar question to the user's summarized question $\mathrm{q_u}$:

\begin{equation}
    \mathrm{q}_i = \argmax_{\mathrm{q} \in \mathcal{Q}} f(\mathrm{q}, \mathrm{q_u})
\end{equation}

\noindent where $\mathcal{Q}$ is the set of questions (FAQs) in the knowledge base, and $f$ is a given similarity scoring function. $\mathrm{q_u}$ is the reference FAQ (during training) or a generated summary (during inference).

We find the set $\mathcal{R}$ of $n$ relevant answer sentences such that it maximizes the relevance score with the user's summarized question $\mathrm{q_u}$:

\begin{equation}
    \mathcal{R} = \argmax_{\mathcal{R'} \subset \mathcal{A}_i} \sum_{\mathrm{a} \in \mathcal{R'}} g(\mathrm{a}, \mathrm{q_u})
\end{equation}

\noindent where $\mathrm{a}$ is an answer sentence, and $g$ is a given relevance scoring function.

\section{Our Pipeline}

Our proposed pipeline for Consumer Health Question Understanding and Answering has three main components.

In the first step, our approach learns to \textit{understand} the intent of user questions (CHQs) by summarizing them. We use an encoder-decoder-based summarization model for this step.

The second step is question matching, or the retrieval of the relevant FAQ from the knowledge base: we \textit{ground} the generated summary to a medical knowledge base of FAQs and corresponding answer documents. As there are no question matching labels, we consider semantic similarity as a proxy to question matching, and we optimize a self-supervised similarity loss.

The third step is the retrieval of the relevant answer sentences: our model learns to \textit{select} the top-$k$ most relevant answer sentences from the matching answer document. To achieve this task in the absence of answer relevance labels, we consider semantic similarity as a proxy for relevance, and we optimize two novel, semantically-guided, and self-supervised loss functions. The first pushes the model to discriminate between relevant and irrelevant sentences, and the other pushes the model to consider only a fixed number of sentences as relevant.

We show an overview of the model and learning objectives in Figure \ref{summary}. The entire pipeline is trained together, as the summarizer encoder is re-used to encode the questions and answer sentences.

\subsection{Question Understanding through Summarization}
\label{question_und}

\begin{figure*}
    \centering
    \includegraphics[width=410pt]{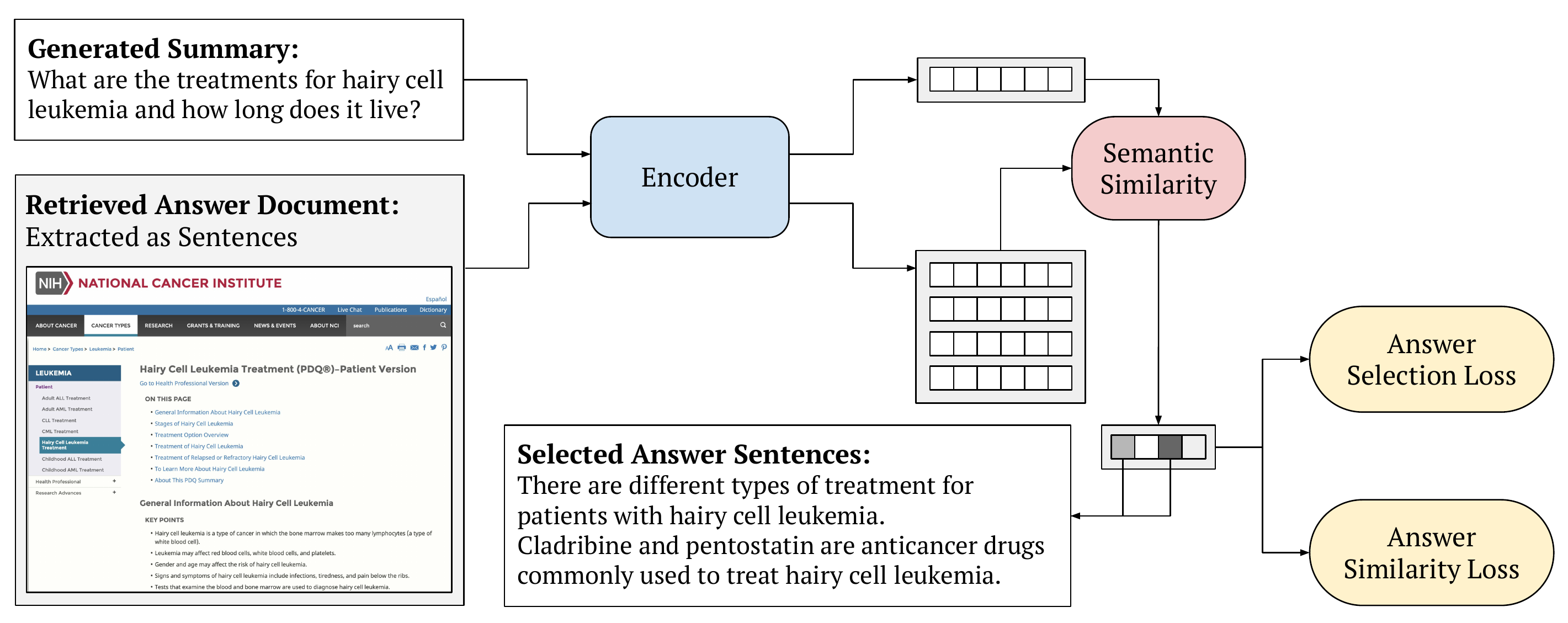}
    \caption{Illustration of the third step of our pipeline: answer retrieval through self-supervised similarity and selection losses (\S \ref{answer_ret}). Following the same example as Figure \ref{knowledge}, our model encodes sentences from the retrieved answer document from the knowledge base, and compares them to the FAQ generated by the summarization model. We use the encoder of the summarization model to embed sentences.}
    \label{selection}
    
\end{figure*}

Our work aims to flip the burden of question understanding on the question answering model. Instead of asking the user to shorten or reformulate their question, we train an encoder-decoder abstractive summarizer to shorten user questions. Figure \ref{knowledge} illustrates this part of the model.

At training time, we input a Consumer Health Question (CHQ) to the summarization model. The reference Frequently Asked Question (FAQ) is the corresponding shorter and formal question. Given a CHQ embedding $\mathbf{x}$ and the corresponding reference FAQ embedding $\mathbf{y_{ref}}$, the summarization loss is defined as the following negative log-likelihood objective:

\begin{equation}
     \mathcal{L}_{\mathrm{sum}} = -\textup{log}p(\mathbf{y}_\mathrm{ref}|\mathbf{x};\theta)
\label{eq1}
\end{equation}

\subsection{Question Matching through Self-Supervised Knowledge Grounding}
\label{question_ret}

In the next step, we match the summarized user question with the most relevant FAQ from the medical knowledge base. We use semantic similarity as a proxy for question matching, in the absence of such labels. 

The knowledge-grounding process is comprised of two steps. First, we use TF-IDF-weighted bag-of-word and $n$-gram vectors to get the top $k$ most relevant FAQs from the knowledge base. This first step acts as a fast filter to extract a small subset of candidate FAQs. Our retrieval approach follows the retrieval methods commonly used in question answering systems \cite{chen2017reading, dinan2018wizard}. \citet{dinan2018wizard} note that the retriever is a potentially learnable part of the model. In our case, using TF-IDF retrieval is computationally optimal and scalable given a large knowledge base with thousands of FAQs. We use a TF-IDF embedder fitted on all the FAQs of the knowledge base, as well as reference FAQs from the training set of the question summarization dataset.

The second step of knowledge-grounding is to rank the top $k$ FAQs using semantic similarity. To get semantic embeddings of the generated summary and the corresponding top $k$ most relevant FAQs from the knowledge base, we use the encoder of the summarization model. We take inspiration from the precision formula of BERT\textsc{Score} \cite{zhang2019bertscore}, and compute the weighted semantic similarity score as follows:

\begin{equation}
\small
     \textup{Sim}(\mathrm{q_u}, \mathrm{q}_i) = \sum_{w \in \mathcal{W}_\mathrm{u}} \max_{w' \in \mathcal{W}_i} \frac{ \mathrm{idf}(w) \cdot \mathrm{CosSim}(\mathbf{x}_w, \mathbf{x}_{w'})}{\sum_{w'' \in \mathcal{W}_\mathrm{u}} \mathrm{idf}(w'')}
\label{semsim}
\end{equation}

\noindent where $\mathrm{q_u}$ is the reference FAQ (during training) or the generated summary (during inference), $\mathrm{q}_i$ is the $i$-th question from the top $k$ most relevant FAQs, $\mathcal{W}_\mathrm{u}$ and $\mathcal{W}_i$ are the corresponding sets of words, $\mathrm{CosSim}$ is the cosine similarity function, and $\mathrm{idf}(w)$ is the inverse document frequency of the word $w$.

The matching FAQ is the knowledge base FAQ with the highest similarity score with  $\mathrm{q_u}$, as shown in the example in Figure \ref{knowledge}. During training, the summarization model may produce low-quality or degenerate FAQs. For this reason, at training time, we choose to use the reference FAQ instead to compute the semantic similarity scores and find the matching FAQ. At test time, we only use the generated summary.

Since we are using different datasets for the question summarization and for the knowledge base, we have to reconcile the questions from the knowledge base and the reference questions. We propose to force the model to learn a representation space that does not distinguish between the reference FAQ and the most similar knowledge base FAQ. To accomplish this, we compute the matching FAQ similarity loss. Given the embedding of a summarization reference FAQ $\mathrm{q}_\mathrm{sum}$ and the embedding of a matching FAQ $\mathrm{q}_\mathrm{mat}$, the matching FAQ similarity loss is defined as:

\begin{equation}
    \mathcal{L}_{\mathrm{mat}} = 1 - \mathrm{ReLU}\left(\mathrm{Sim}\left(\mathrm{q}_\mathrm{sum}, \mathrm{q}_\mathrm{mat}; \theta \right) \right )
\label{retrieved_faq}
\end{equation}

\subsection{Answer Retrieval through Self-Supervised Similarity and Selection Losses}
\label{answer_ret}

After summarizing the user question and retrieving a relevant FAQ from the knowledge base, the next step is to retrieve relevant sentences from the corresponding answer document. In our setting, we need to retrieve a fixed number of sentences relevant to the user question. However, we have no labels for the answer sentences indicating relevance to the user question. We propose two complementary self-supervised learning objectives, that use semantic similarity as a proxy to relevance scoring, and satisfy the constraint of selecting a fixed number of answer sentences.

We show an overview of our answer retrieval approach in Figure \ref{selection}. In the example of the figure, we show for simplicity a relatively short answer document with four sentences, from which the model chooses the two most relevant ones. In practice, there are close to ten sentences in answer documents.

We compute semantic similarity scores between the generated summary (for inference) or the reference FAQ (for training), and each of the sentences of the retrieved answer document. We obtain the semantic embeddings of each sentence using the encoder of the summarization model. We then compute semantic similarity scores as shown in equation \ref{semsim}. Cosine similarity scores have values in the $[-1;1]$ range. For a pair of sentences, a cosine similarity value closer to $-1$ means that the corresponding sentence embeddings are negatively correlated, or that the sentences have opposite meanings. A value closer to $0$ means that the embeddings are not correlated, and that there is no particular semantic relation between the sentences. A value closer to $1$ means that the sentence embeddings are positively correlated, and the sentences are close semantically. We consider that a sentence is relevant when the values are closer to $1$, and irrelevant otherwise. For this reason, we apply a $\mathrm{ReLU}$ activation on the cosine similarity scores before feeding them to the loss functions.

We propose two learning objectives to achieve the self-supervised selection of relevant answer sentences. The semantic similarity loss pushes the model to increase its confidence in the relevance of answer sentences, whereas the answer selection loss pushes the model to select only a fixed number of sentences. The intuition for sharing the encoder with the summarization model, is that these two losses will enable the summarizer to absorb notions of relevance and semantic similarity.

Given the summarization reference FAQ $\mathrm{q_{sum}}$ and the $i$-th sentence of the retrieved answer document $\mathrm{a}_i$, we compute the $\mathrm{ReLU}$-activated semantic similarity score as follows:

\begin{equation}
    S(\mathrm{q_{sum}}, \mathrm{a}_i; \theta) = \mathrm{ReLU}\left(\mathrm{Sim}\left(\mathrm{q_{sum}}, \mathrm{a}_i; \theta \right) \right )
\label{relusim}
\end{equation}

We then define the semantic similarity loss $\mathcal{L}_{\mathrm{sim}}$ and the answer selection loss $\mathcal{L}_{\mathrm{sel}}$ as follows:

\begin{equation}
    \mathcal{L}_{\mathrm{sim}} = \sum_{i = 1}^{|\mathcal{A}|} S(\mathrm{q_{sum}}, \mathrm{a}_i; \theta) * \left(1 - S(\mathrm{q_{sum}}, \mathrm{a}_i; \theta)\right)
\label{semsimloss}
\end{equation}

\begin{equation}
    \mathcal{L}_{\mathrm{sel}} = \left|\mathrm{min}(n,|\mathcal{A}|) - \sum_{i = 1}^{|\mathcal{A}|} S(\mathrm{q_{sum}}, \mathrm{a}_i; \theta)\right|
\label{ansselloss}
\end{equation}

\noindent where $\mathcal{A}$ is the set of sentences in the retrieved answer document, and $n$ is the fixed number of sentences to be retrieved.

The semantic similarity loss $\mathcal{L}_{\mathrm{sim}}$ pushes the semantic similarity values to be either $1$ (relevant) or $0$ (irrelevant). In combination with $\mathcal{L}_{\mathrm{sim}}$, the answer selection loss pushes the model to only select up to $n$ sentences to have semantic similarity values close to $1$. Our system then outputs the sentences with the highest semantic similarity values in the order in which they appear in the answer document. Therefore, the particular semantic similarity ranking of the relevant sentences does not matter -- it only matters that relevant sentences have the $n$ highest values.

Finally, the learning objective $\mathcal{L}$ is as follows:

\begin{equation}
    \mathcal{L} = \mathcal{L}_{\mathrm{sum}} + \lambda * \mathcal{L}_{\mathrm{mat}} + \gamma * (\mathcal{L}_{\mathrm{sim}} + \mathcal{L}_{\mathrm{sel}})
\label{loss}
\end{equation}

\noindent where $\lambda$ and $\gamma$ are hyperparameters. We use only one weight for $\mathcal{L}_{\mathrm{sim}}$ and $\mathcal{L}_{\mathrm{sel}}$ as these two losses are complementary.

\section{Experiments and Results}

In this section, we evaluate our proposed pipeline for Consumer Health Question Understanding and Answering, and we propose to compare our proposed pipeline against two strong baselines. Seven medical experts judge the performance of our system and baselines by asking their own questions, and rating the relevance of the answers retrieved. Then, we analyze the results through the lens of summarization metrics, human evaluation, and computational speed.

\subsection{Datasets}

We use one medical knowledge base, MedQuAD \cite{abacha2019question}, and two medical question summarization datasets: MeQSum \cite{abacha2019summarization} and HealthCareMagic \cite{chen2020meddiag}. All datasets are in English. We show dataset statistics in Table \ref{datastats}.

\begin{table}[]
    \centering
    \begin{tabular}{|l|r|r|r|}
    \hline
        \sc Dataset split & \sc Train & \sc Dev & \sc Test  \\ \hline
         MeQSum & 405 & 50 & 50 \\ \hline
         HealthCareMagic & 1,314 & 164 & 165 \\ \hline
    \end{tabular}
    \caption{Statistics of the medical dataset splits.}
    \label{datastats}
\end{table}

\subsubsection{Dataset Details}

\textbf{MedQuAD} is a large-scale Medical Question Answering Dataset. \citet{abacha2019question} collect trusted medical question-answer pairs by crawling them from 12 websites of the U.S. National Institutes of Health (NIH). Each web page contains information about a health-related topic, like a disease or a drug. The authors automatically collect the question-answer pairs by composing handcrafted patterns adapted to each website based on document structure and section titles. They manually evaluate 1,721 CHQs to come up with automatic wording patterns for each of 36 question types. Therefore, even though answers are curated and written by medical experts, questions are automatically formulated and may have some noise.

We collect the publicly available (e.g. not copyrighted) question-answer pairs from the MedQuAD dataset\footnote{\url{https://github.com/abachaa/MedQuAD}}. We then use the NLTK sentence tokenizer \cite{bird2006nltk} to split answer documents into sentences. We get 16,423 questions and 157,592 answer sentences, making for an average of 9.6 answer sentences for each question.

\textbf{MeQSum} \cite{abacha2019summarization} is a medical question summarization dataset released by the U.S. National Institutes of Health (NIH). It contains 1,000 consumer health questions summarized into FAQ-style single-sentence questions by medical experts.

\textbf{HealthCareMagic} is a medical dialogue dataset issued as part of the MedDialog dataset \cite{chen2020meddiag}\footnote{\url{https://github.com/UCSD-AI4H/Medical-Dialogue-System}}. It is crawled from \texttt{HealthCareMagic.com}, an online healthcare service platform. This dataset includes first a formally worded, one-sentence question describing the intent of the patient question, followed by 2 long utterances: a CHQ from the patient that includes a description of the problem and a question, and then an answer from the doctor. To form a medical question summarization dataset, we consider the single-sentence descriptions as summaries of the patient's CHQ. We collect 226,405 question pairs.

\subsubsection{Knowledge-based Filtering of Datasets}

We conduct experiments for each of the two question summarization datasets, and we use MedQuAD as the underlying knowledge base in all experiments. For this reason, we decide to filter each of the question summarization datasets to reconcile their differences with MedQuAD.

We first fit a TF-IDF embedding model, similar to the one of \cite{dinan2018wizard}, on the reference FAQs of each question summarization dataset and the questions of MedQuAD. We then compute the dot products of the TF-IDF-weighted vectors for all possible pairs of summarization FAQs and MedQuAD questions. We assign a matching score $m(\mathrm{q_{sum}})$ to each summarization reference FAQ:

\begin{equation}
    m(\mathrm{q_{sum}}) = \max_{\mathrm{q'} \in \mathcal{Q}_\mathrm{MedQuAD}} \mathrm{tfidf}(\mathrm{q_{sum}}) \cdot \mathrm{tfidf}(\mathrm{q'})
\end{equation}

We manually evaluate the matching scores for each summarization dataset to set a cutoff matching score of filtering. This way, we obtain question summarization datasets where reference FAQs have matches in the medical knowledge base. Finally, we perform a random and rough 80/10/10 split for the train/dev/test sets. The dataset statistics are in the main paper.

\subsection{Training Settings}

We adopt the BART encoder-decoder model \cite{lewis2019bart}, as it set a state of the art in abstractive summarization benchmarks. We train our model using the HuggingFace implementation \cite{wolf2020transformers}, on a learning rate of $2 \cdot 10^{-6}$. The question matching pool retrieved by TF-IDF is comprised of $k = 32$ knowledge base FAQs. Our answer selection loss $\mathcal{L}_\mathrm{sel}$ is optimized to select up to $n = 3$ sentences. We use $\lambda = 0.01$ and $\gamma = 0.01$ as weights for the self-supervised losses. The BART encoder is used for embedding sentences for question matching and answer selection.

We train for 50 epochs for MeQSum, and 20 epochs for HealthCareMagic. Each training epoch takes about 10 minutes for MeQSum, and about 35 minutes for HealthCareMagic. Inference takes 1 minute for the MeQSum test set and 3 minutes for the HealthCareMagic test set. The best checkpoint is selected based on the lowest loss value $\mathcal{L}$ on the dev set.

We use BART Large pre-trained on the CNN-Dailymail dataset, and each BART Large model contains 406 million parameters, as per the HuggingFace implementation.

\begin{table*}[]
    \centering
    \begin{tabular}{|l|r|r|r|}
        \hline
        \sc System & MeQSum & HealthCareMagic & Time/Query \\ \hline
        DPR \cite{karpukhin-etal-2020-dense} & 1.42 & 1.73 & 47 seconds \\ \hline
        GAR \cite{mao2021generation} & 1.40 & 1.64 & 48 seconds \\ \hline
        Ours & 2.13 & 2.35 & 2 seconds \\ \hline
    \end{tabular}
    \caption{Evaluation of the relevance (out of 5) of answers retrieved by our proposed system and two strong baselines for questions asked by seven evaluators. The systems trained on MeQSum are evaluated on 60 questions by 3 evaluators, and the ones trained on the larger HealthCareMagic dataset are evaluated on 80 questions by 4 evaluators. The column on the right shows the number of seconds it takes for a loaded system to retrieve the answer to a query.}
    \label{evaluation}
\end{table*}

\subsection{Baselines}

We propose the two following baselines in retrieval-based question answering: Dense Passage Retrieval (DPR) \cite{karpukhin-etal-2020-dense}, and Generation-Augmented Retrieval (GAR) \cite{mao2021generation}. We adapt these two baselines to our case, and adopt BART-based pre-trained encoders.

Similarly to our own pipeline, we create a two-stage retrieval to get answers. The first stage encodes questions from the knowledge base, and retrieves the question that is most relevant to the query. The second stage encodes the corresponding answer document, and retrieves the three sentences that are most relevant to the query.

For DPR, the query is simply the user question. For GAR, we need to generate a context to add to the user question: we choose to add the summary of the user question as the context. We train a BART encoder to summarize user question, using the question summarization datasets.

Whereas our system's retrieval encoder is trained on our proposed self-supervised objectives, the retrieval encoders of the baselines are trained on Wikipedia for the task of retrieval-based question answering.

\subsection{Do we retrieve relevant answers?}

\subsubsection{Evaluation Strategy}

We hire seven annotators: four of which are medical doctors, and the remaining three hold degrees related to healthcare or immunology.

We ask the evaluators to first write user questions, and then evaluate the answers retrieved by our system and the two existing systems. Given that our medical knowledge base has limited questions, we ask the evaluators to limit their questions to the topics covered by the nine sources from which the knowledge base was extracted.

Then, we ask the evaluators to rate the relevance of the answers retrieved by each system independently, on a scale of 1 (not relevant) to 5 (relevant). The full description of scores given to the annotators is in the Appendix.

Each of the seven annotators wrote 20 questions, and each question gets three answers (one per system). We assign three annotators to the models trained on MeQSum, and four to the models trained on HealthCareMagic. The annotators rate answers only for the questions that they wrote themselves.

\subsubsection{Results and Discussion}

\begin{table*}[]
    \centering
    \begin{tabular}{|l|r|r|r|r|r|r|r|r|r|r|r|r|}
    \hline
        \sc Criteria & \multicolumn{3}{|c|}{Fluency} & \multicolumn{3}{|c|}{Coherence}  & \multicolumn{3}{|c|}{Informativeness}  & \multicolumn{3}{|c|}{Correctness}  \\ \hline
        \sc Evaluation & Win & Lose & Tie &  Win & Lose & Tie &  Win & Lose & Tie &  Win & Lose & Tie \\ \hline
        MeQSum & 11 & 5 & 28 & 10 & 6 & 28 & 12 & 3 & 29 & 12 & 4 & 28 \\ \hline
        HealthCareMagic & 45 & 17 & 42 & 44 & 19 & 41 & 46 & 18 & 40 & 44 & 18 & 42 \\ \hline
    \end{tabular}
    \caption{Question Understanding evaluation: blind evaluation by 2 annotators of the generated summaries for the test set CHQs. A ``Win'' evaluation means that our model generates a better summary than the baseline summarizer.}
    \label{evalsum}
    
\end{table*}

\begin{table}[]
    \centering
    \resizebox{\columnwidth}{!}{
    \begin{tabular}{|p{3.9cm}|r|r|r|r|r|r|}
    \hline
        \sc Dataset & \multicolumn{3}{|c|}{MeQSum} & \multicolumn{3}{|c|}{HealthCareMagic} \\ \hline
        \sc Metric & R1 & R2 & RL & R1 & R2 & RL \\ \hline
        GAR \cite{mao2021generation} & 45.72 & 30.43 & 42.02 & 31.04 & 13.68 & 27.90 \\ \hline
        Ours & 46.74 & 30.10 & 42.81 & 33.13 & 14.71 & 30.18 \\ \hline
    \end{tabular}}
    \caption{Question Understanding evaluation: summarization results on test set (reference FAQs). The R1, R2 and RL metrics refer to the F1 scores of \textsc{Rouge}-1, \textsc{Rouge}-2 and \textsc{Rouge}-L.}
    \label{rouge}
    
\end{table}

We show the results of the evaluations in Table \ref{evaluation}. The first three columns show the averages of relevance scores that were given by annotators for all systems.

The results show that the evaluators have preferred our system's answers over the answers retrieved by the two baselines. Our system gets relevance scores that are 0.6 to 0.7 points higher, out of 5 on the relevance scale. An annotator commented that they find our system to be "\textit{more organized and to-the-point than the rest of systems}."\footnote{Annotators were not told that either system was ours or not. The systems were simply numbered for a blind evaluation.}

The two baselines seem to perform similarly to each other. This is likely due to the fact that the main difference between them is that the query is generation-augmented for GAR, whereas the query is simply the user question for DPR.

Overall, the relevance scores are on the lower side, as no system exceeds an average score of 2.5/5. This shows that consumer health question answering and understanding is a challenging task, especially since there are no labels to indicate whether an answer is relevant to a particular question, or which FAQ matches the user's intent.

In addition, the challenges of the task are also due to the limitations of the knowledge base. Some annotators noted that the retrieved answers were often not appropriate, or close to the topic but not answering the question. This is due to the fact that MedQuAD does not cover all possible illnesses and medical conditions that the users could ask about. Whereas a larger database would potentially solve coverage problems, it could be at the expense of the quality or verifiability of the answers. The MedQuAD dataset is at times noisy, and contains generic sentences that may not answer any question, or generic templates related to percentages of symptoms and how frequent they are.

\subsection{Computational Speed}

We run our system on a single 11GB GPU, whereas the two baselines are each run on four 16GB GPUs. We show the average duration required to retrieve answers for a single query in the right column of Table \ref{evaluation}.  

We notice that, in addition to the higher relevance scores, the advantage of our system is that it is significantly (more than 20 times) faster compared to the two baselines. This is largely due to the fact that we limit to 32 the number of knowledge base questions that we encode and compare the query embedding to. In contrast, DPR and GAR encode all questions in the knowledge base. This is done at the beginning when loading the models, but the query similarity computation is done at each run, thereby lengthening the processing time.

\subsection{Analysis of Question Understanding}

An additional way that our system outperforms the two baselines could be through summarization. We evaluate the summarization of consumer health questions using the \textsc{Rouge} metric \cite{lin2004rouge}. Our GAR baseline uses a BART model trained on the summarization loss only. We show the results in Table \ref{rouge}. We notice that sharing encoder parameters between the summarization loss and our proposed self-supervised losses generally increases \textsc{Rouge} F1 scores across both datasets. For HealthCareMagic, score increases exceed 2 points in \textsc{Rouge}-1 and \textsc{Rouge}-L.

Given that \textsc{Rouge} is notoriously unreliable, we hire two additional annotators on Upwork who are healthcare workers to judge the fluency, coherence, informativeness and correctness of generated summaries. We show the annotators the consumer health question (source text), the reference FAQ (target text) and two generated summaries. The annotators do not know which system generated which summary. We show the evaluation scores in Table \ref{evalsum}. We remove repetitions of reference FAQs in the test sets put up for evaluation. The results confirm that our self-supervised losses increase the quality of generated summaries. Summaries generated with our model score more wins more often than losses on all four metrics, and score more wins than ties with the summarization-only baseline for HealthCareMagic.

\section{Conclusions}

We introduce an end-to-end pipeline for knowledge-grounded consumer health question answering and understanding (CHQUA). Our challenge is that we have no labels for question matching or answer relevance. We propose to use semantic similarity as a proxy for those labels, and we design three novel self-supervised losses: one works to match the user's summarized question to a knowledge base question, and the other two losses work complementarily to teach our model to select a fixed number of relevant answer sentences. 
We compare our proposed system against two strong baselines of retrieval-based question answering. We hire seven medical experts to ask their questions, and they find that our system provides more relevant answers. Our system also achieves processing times that are more than 20 times faster. Finally, we find that our proposed self-supervised losses enable the summarizer model to achieve higher scores in ROUGE and human evaluation metrics, compared to a summarization-only baseline. However, we find that this task remains challenging and that there is still room for improvement. We release our code and model to encourage further research.

\section*{Ethical Considerations}

Our model is for medical question answering, but should be used with caution as it does not claim to provide medical advice. Potential users of our system should be warned to not blindly trust the answers given to their medical questions. Potential users should always consult their physician for medical advice.

Each of our annotators spent between two and four hours on the task we gave them. Each annotator was compensated fairly for their work. We answered all of the annotators' questions about the task before they started.  Hiring platform Upwork guarantees the payment, fair treatment and informed consent of our nine hired annotators through a mutually agreed-upon contract. The platform fee for Upwork was paid by us, and not deducted from the compensation of the annotators.

\section*{Acknowledgements}

Khalil Mrini is supported by Adobe Research Unrestricted Gifts, and by an Amazon Research Award under his AWS AI proposal ``\textit{Learning Representations for Voice-Based Conversational Agents for Older Adults}''. This work is part of the VOLI project \cite{khalilmedical, johnson2020voice}, and we gratefully acknowledge the award from NIH/NIA grant R56AG067393.

\bibliography{anthology,custom}
\bibliographystyle{acl_natbib}

\pagebreak

\appendix

\section{Annotation Details}

\subsection{Topics Covered by the Knowledge Base}

We ask the annotators to limit their questions to the nine sources of MedQUAD. The nine sources from which questions and answer documents are extracted are as follows:

\begin{itemize}
    \item National Cancer Institute
\item Genetic and Rare Diseases Information Center: various aspects of genetic/rare diseases
\item Genetics Home Reference (GHR): consumer-oriented information about the effects of genetic variation on human health
\item MedlinePlus Health Topics: information on symptoms, causes, treatment and prevention for diseases, health conditions, and wellness issues
\item National Institute of Diabetes and Digestive and Kidney Diseases
\item National Institute of Neurological Disorders and Stroke: neurological and stroke-related diseases 
\item NIHSeniorHealth: health and wellness information for older adults
\item National Heart, Lung, and Blood Institute (NHLBI): diseases, tests, procedures, and other relevant topics on disorders of heart, lung, blood, and sleep
\item Centers for Disease Control and Prevention (CDC)

\end{itemize}

\subsection{Answer Relevance Scoring}

We ask annotators to rate answers retrieved by our system and the two baselines according to the following criteria:

\begin{itemize}
    \item Score of 1/5: The system’s answer is completely irrelevant to the question, and does not even contain any concept related to the question.
\item Score of 2/5: The system’s answer mentions notions that are related to the question, but does not contain a word or concept mentioned in the question.
\item Score of 3/5: The system’s answer mentions one or more words or concepts from the question, but does not actually answer the question.
\item Score of 4/5: The system’s answer partially answers the question, mentions one or more words or concepts from the question, but does not fully answer the question.
\item Score of 5/5: The system’s answer fully answers the question.
\end{itemize}

\subsection{Question Understanding}

For question summarization, we evaluate the generated summaries on 4 criteria. We define these criteria for the two healthcare worker annotators as follows:

\begin{itemize}
    \item Fluency: which generated FAQ is more grammatically correct, and easier to read and to understand?
    \item Coherence: which generated FAQ is better structured and more organized?
    \item Informativeness: which generated FAQ captures the most out of the concern of the patient who wrote the CHQ?
    \item Correctness: which generated FAQ is more factually correct given the CHQ?
\end{itemize}

\subsection{Upwork}

\begin{figure*}
    \centering
    \includegraphics[width=\textwidth]{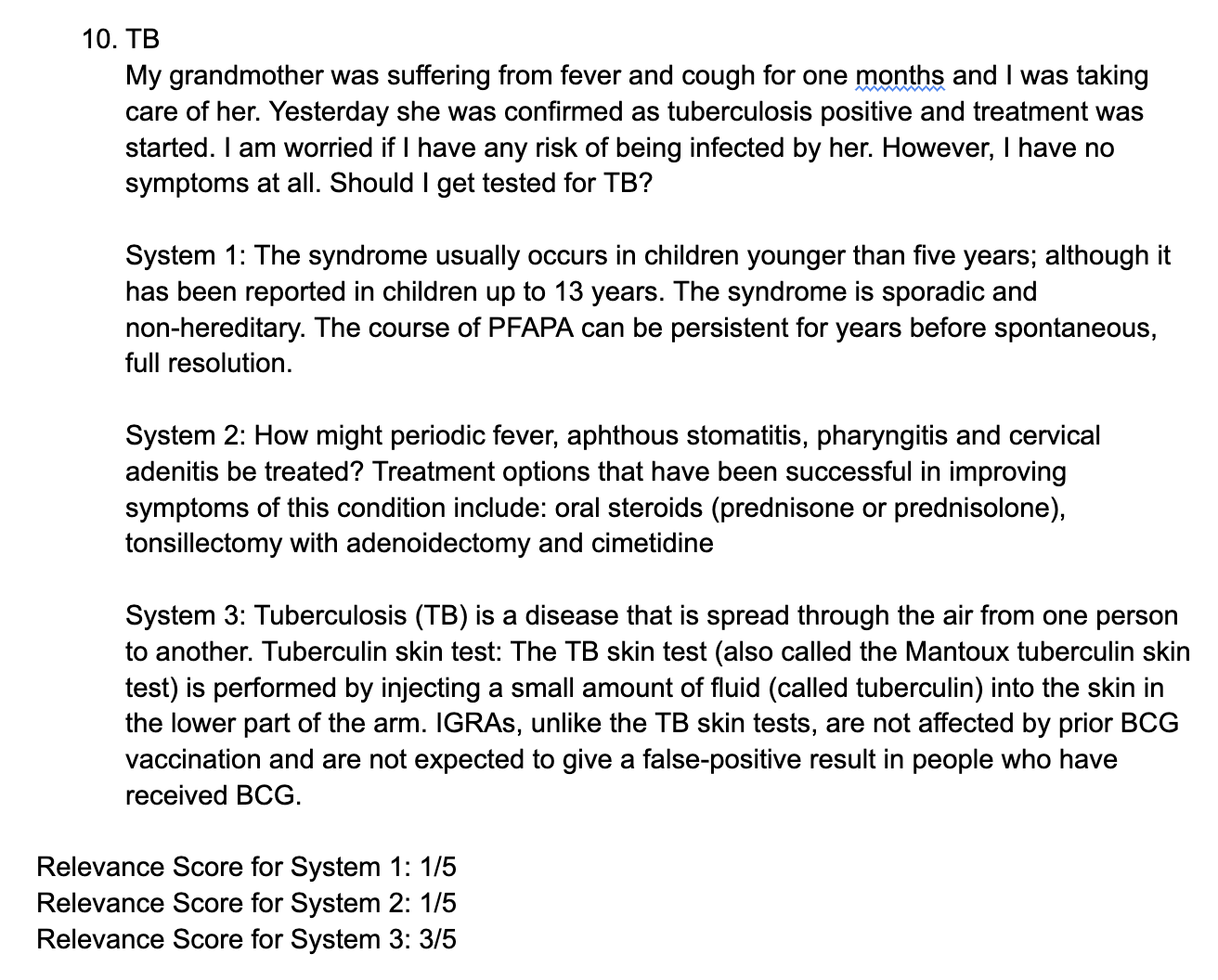}
    \caption{Example of a Google document, where a hired annotator (medical doctor) asks a question, and rates the answers that we pasted once retrieved by our system and the two baselines.}
    \label{interface}
\end{figure*}

We ask annotators to work on Google docs that we share with them. We show in Figure \ref{interface} an example of a Google doc that we shared with an annotator (medical doctor) to ask their own question, and the answers we pasted for them to evaluate.

\end{document}